\documentclass[11pt]{article}
\usepackage[includeheadfoot,margin=1in]{geometry}
\usepackage{times}
\usepackage{cite}
\usepackage{amsmath,amsfonts,amssymb, booktabs,url}
\usepackage{mathtools}
\usepackage{enumerate}
\usepackage{verbatim}
\usepackage{commath}
\usepackage{graphicx}
\usepackage{booktabs}
\usepackage{tabularx}
\DeclareUnicodeCharacter{2013}{-}
    \usepackage[usenames]{color}
    \definecolor{plum}  {rgb}{.4,0,.4}
    \definecolor{BrickRed} {rgb}{0.6,0,0}
	\definecolor{DarkBlue} {rgb}{0,0,0.6}
\usepackage[plainpages=false,pdfpagelabels,colorlinks=true,linkcolor=BrickRed,citecolor=plum,urlcolor=DarkBlue]{hyperref}
\usepackage[mathcal]{euscript}
\usepackage{cleveref}
\begin{document}
\title{\LARGE Accelerating Divisible Load Processing Through Machine Learning: A Practical Framework for Large-Scale Workloads}
\author{\Large Bharadwaj Veeravalli \\ \href{mailto:elebv@nus.edu.sg}{elebv@nus.edu.sg} \\ %
Department of Electrical and Computer Engineering,\\ National University of Singapore, 4 Engineering Drive 3, Singapore.}
\date{\today}
\maketitle
\begin{abstract} 
In this paper, we introduce the first machine learning framework for predicting optimal processing times in Single-Level Tree Network (SLTN) architectures for the Divisible Load Theory (DLT) paradigm. Using a feedforward neural network(FNN) with 16 engineered features, we train a model on 100,000 synthetically generated configurations to predict optimal processing times without explicit formulation of DLT
equations. The model achieves 97-99\% accuracy (R²) with mean absolute percentage error of 1-5\%, demonstrating that neural networks can effectively learn complex load distribution relationships. Feature importance analysis reveals that the model implicitly captures DLT
mathematical structure, including load conservation and simultaneous finishing constraints. With inference times under 1 millisecond, the approach serves as a viable option over traditional DLT computation, enabling applications in real-time scheduling, design space exploration,
and cloud resource allocation. The method generalizes well across diverse system configurations (n=3-20, load=1-100 GB) with consistent accuracy, though performance degrades slightly for very large or highly heterogeneous systems. This work demonstrates the feasibility of using machine learning to accelerate distributed computing optimization
while maintaining near-optimal accuracy.
\end{abstract}
%
\section{INTRODUCTION}
Given the growing interest and widespread applicability of machine learning across diverse domains, this work aims to introduce a machine learning approach into the Divisible Load Theory (DLT) literature and demonstrate its potential for direct application in various fields. So far, the DLT domain literature has been perceived primarily as a theoretical support framework for handling large-scale workloads. It was used as a tool to determine the ultimate performance bounds and also to derive average performance limits for network-based computing. However, almost all the  strategies proposed in the earlier era demanded explicit theoretical development and programming which was challenging. In current times, with hardware and software advancements, machine learning approach becomes a viable and efficient approach to determine solutions almost instantaneously without involving explicit computational or programming  steps. 

\subsection{Machine learning: A data-driven approach for DLT}
There are several compelling reasons as to why and how machine learning approach seems imperative to adopt at this stage. Traditional DLT requires explicit solution of recursive equations to determine optimal load distribution for each system configuration, necessitating iterative computation of interdependent parameters every time system characteristics change—whether due to variations in system size, processor speeds, link bandwidths, or workload magnitudes. This computational overhead becomes particularly prohibitive in dynamic environments where system parameters fluctuate frequently, such as cloud computing infrastructures experiencing processor failures or replacements, network bandwidth variations due to congestion, or heterogeneous clusters with time-varying compute capabilities. Furthermore, extending DLT to incorporate practical real-world aspects like processor addition/removal during execution, dynamic bandwidth allocation, multi-level hierarchical networks, heterogeneous task requirements, or result collection phases exponentially increases the complexity of analytical solutions, often rendering closed-form optimization intractable. \\[0.3cm]
Machine learning offers a fundamentally different paradigm: by training neural networks on large datasets encompassing diverse static configurations—varying system sizes ($n$), low to high order of compute speeds, link bandwidths and load magnitudes, the model learns to approximate the underlying mathematical relationships between system parameters and optimal processing times without explicitly solving recursive equations at inference time. This data-driven approach enables near-instantaneous predictions (sub-millisecond inference) compared to iterative DLT computation, achieving $10-100$× computational speedup while maintaining $95-99\%$ accuracy, making it particularly valuable for real-time scheduling decisions, rapid design space exploration evaluating thousands of architectural alternatives, online optimization in production systems requiring immediate load distribution decisions, and future extensions to dynamic scenarios where continuous re-optimization would be computationally prohibitive using traditional analytical methods. Moreover, a trained model inherently captures complex non-linear interactions between heterogeneous processor speeds and network bandwidths that might be difficult to express in closed-form equations, providing a pathway toward handling increasingly realistic distributed computing scenarios where analytical tractability becomes a limiting factor for practical deployment.

Classical literature on DLT can be readily accessed in \cite{Bharadwaj1996, Robertazzi2014}. In this section, we focus on the most pertinent works to align readers with the contextual background and to elucidate the motivation for transitioning toward the machine learning domain.\\[0.3cm]
While classical DLT provides closed-form optimal solutions for static configurations, several studies have identified computational limitations when extending to realistic scenarios. Recognizing that real-world distributed systems exhibit uncertainty in network bandwidth, processor speeds, and task characteristics, some earlier works have extended DLT to such uncertain and stochastic settings. The concept of adaptive strategies was first introduced in 2002 \cite{Ghose2002}, with its core mechanism relying on probing real-time network resource availability and dynamically adjusting load distribution using feedback information to optimize scheduling efficiency. This approach proves particularly effective in distributed environments characterized by dynamic resource variations, as it significantly reduces task completion delay while enhancing system performance. Work in \cite{LiK2014} formulated stochastic task scheduling as a linear programming problem that maximizes a weighted objective by combining schedule length and energy efficiency under deadline and energy budget constraints, and proposed a heuristic energy-aware scheduling algorithm. Zomaya and Teh~\cite{Zomaya2001} proposed fuzzy logic-based approaches to DLT where system parameters are represented as fuzzy numbers, enabling graceful handling of imprecise measurements. Work reported in \cite{YYHC2003} explicitly accounts for uncertainty and variability in resource performance. It combines robustness with efficient multi‑round scheduling to maintain good makespan under realistic distributed‑system conditions. Veeravalli et. al.~\cite{BVJY2004} demonstrated DLT scheduling on multi-level tree networks and used spanning tree approaches to derive near-optimal solutions.\\[0.3cm]
Now, with the recent proliferation of IoT devices and edge/fog computing paradigms new DLT research directions were attempted in the recent past. Kazemi et al. ~\cite{Kazemi2020} applied DLT to fog computing architectures with hierarchical processing layers, optimizing workload distribution across edge devices, fog nodes, and cloud servers. Mahmud et al.~\cite{Mahmud2018} developed latency-aware divisible load scheduling for IoT applications with real-time constraints, incorporating end-to-end delay bounds into the optimization framework. Finally, in a very recent work \cite{Gokul2024} DLT has been demonstrated via an experimental implementation on handling large-scale SAR data processing. \\[0.3cm]
With growing concerns about energy consumption in data centers, several studies have integrated energy efficiency into DLT optimization. Drozdowski and Wielebski~\cite{Drozdowski2016} formulated bi-objective optimization problems balancing processing time and energy consumption, demonstrating that slight increases in completion time can yield substantial energy savings. \\[0.3cm]
Although machine learning has been extensively applied to general scheduling problems~\cite{Mao2016ML}, its application to DLT remains largely unexplored. Zhou et al.~\cite{Zhou2018} used reinforcement learning for divisible load scheduling in dynamic environments, learning policies through trial-and-error interactions, but reported slow convergence and difficulty generalizing across different system sizes. To the best of our knowledge, the presented work in this paper is the first to apply {\it supervised learning} with feedforward neural networks to predict DLT optimal processing times directly from system parameters, achieving near-optimal accuracy with orders-of-magnitude speedup compared to analytical computation. 
\subsection{Objectives and Scope of this Work}
The primary objective of this work is to develop and demonstrate the feasibility of a machine learning approach for predicting optimal processing times in Single-Level Tree Network (SLTN) using DLT paradigm without explicit computation of the underlying mathematical equations. Specifically, we aim to: (1) design a feedforward neural network architecture that learns the complex non-linear relationships between system parameters (processor speeds, link bandwidths, system size, and 
load magnitude) and optimal processing times; (2) demonstrate that the trained model achieves a high prediction accuracy (targeting at least $R^2 > 0.90$) with mean absolute percentage errors below 8\% to 10\%; (3) achieve low inference times compared 
to traditional DLT computation, enabling real-time optimization and rapid design space exploration; (4) provide comprehensive explainability analysis showing how 
the neural network implicitly learns DLT mathematical structure (load conservation, simultaneous finishing constraints, and optimal load distribution relationships); 
and (5) validate the approach across diverse system configurations spanning different system sizes ($n=3-20$), heterogeneity levels, and load magnitudes to 
ensure robust generalization. 

The main purpose of this research is to bridge machine learning and distributed computing optimization, opening pathways for ML-accelerated solutions in resource allocation, cloud scheduling, and hardware procurement decisions. Finally, it may be noted that this work is scoped to Single-Level Tree Networks (SLTN) under the standard DLT framework with homogeneous divisible loads and sequential communication from root to child processors.  The model predicts optimal processing times (T*) for static configurations, excluding result collection, multi-level hierarchies, dynamic 
failures, and heterogeneous tasks, with future extensions planned for these aspects.\\[0.3cm]
The organization of this paper is as follows. In Section 2, we present an abridged overview of the classical SLTN results using DLT paradigm. In Sections 3 and 4, we describe the neural network architecture used and the learning process. In Section 5, we present all our performance evaluations with rigorous discussions and conclude the paper with insights and recommendations in Section 6. 
\section{QUICK REVIEW ON SINGLE-LEVEL TREE NETWORK RESULTS}
In this section, we will consolidate and present the results for a single -level tree network from the  literature comprising a root processor and $n$ children nodes.
\subsection{System Model}
We consider a single-level tree network (SLTN) consisting  of a root processor $P_0$ connected to $n$ child processors $P_1, \dots, P_n$, where computation time per unit load on processor $i$ is denoted $w_i$ and communication time per unit load from root to child $i$ is $z_i$. Communication occurs sequentially (single-port) at the root, while all processors including the root have front-ends enabling computation during transmission. The total normalized divisible load equals 1, distributed as fractions $\alpha_i$ to each $P_i$.
\begin{figure}
    \centering
    \includegraphics[width=1\linewidth]{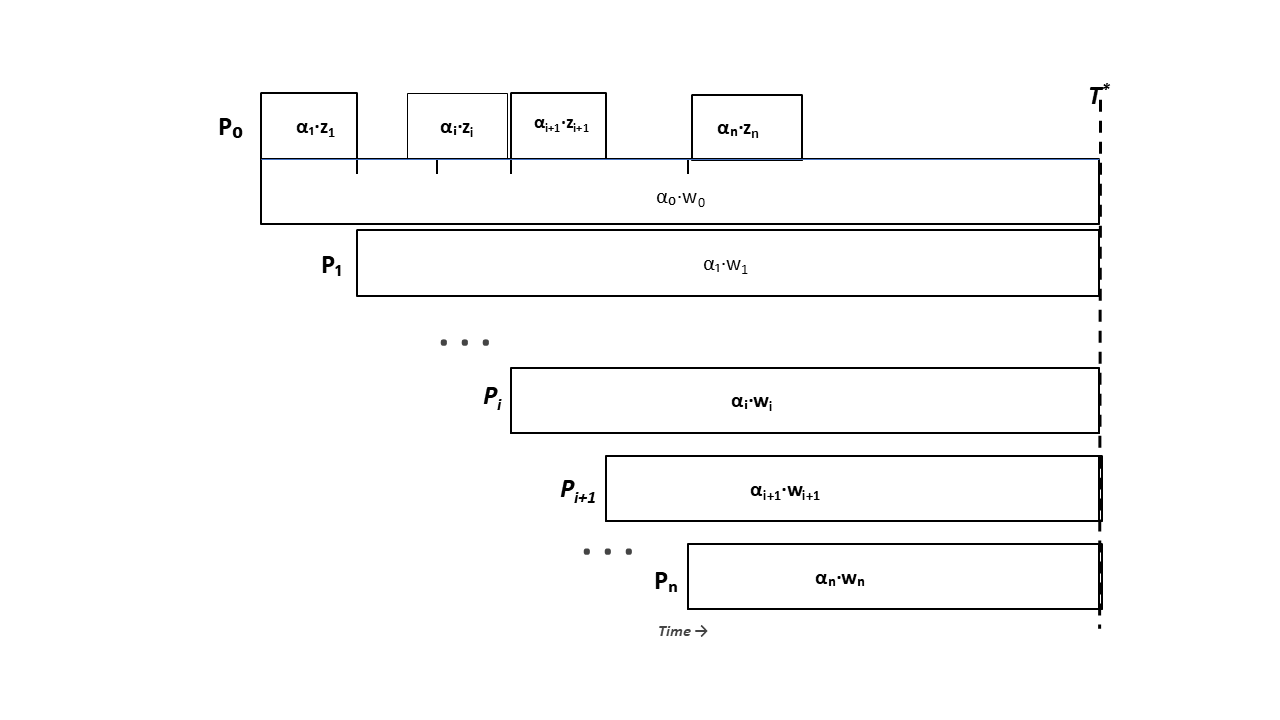}
     \caption{Timing Diagram: Optimal Load Distribution in a Single-level Tree}
    \label{fig:TD}
\end{figure}

\subsection{Timeline and Optimality Conditions}
Fig. \ref{fig:TD} shows a timing diagram to capture the load distribution process.The root $P_0$ divides and starts to distribute respective parts {$\alpha_1,...,\alpha_n$} of the workload $L$ to child processors and also processes its share $\alpha_0$ in a time-overlapped fashion as shown. This is supported by offloading the communication part to a dedicated front-end. The speeds of the processor $P_i$ and link $l_i$ are denoted as, $w_i$ and $z_i$(both (expressed as secs/unit load), respectively, as in the DLT literature. 
Thus, the communication to $P_i$ is given by:
\[
C_i = \sum_{k=1}^{i} \alpha_k z_k.
\]
A child processor $i$ begins computing at time $C_i$ and finishes at:
\[
T_i = C_i + \alpha_i w_i \quad (i=1,\dots,n)
\]
The root node completes its workload $\alpha_0$ at 
\[
T_0 = \alpha_0 w_0
\]
\subsection{Deriving the Fundamental Recursion and an  Optimal solution}
From the Fig. \ref{fig:TD} , for a child $i$:
\[
T_i = \sum_{k=1}^{i} \alpha_k z_k + \alpha_i w_i
\]

Similarly, for a child $(i-1)$:
\[
T_{i-1} = \sum_{k=1}^{i-1} \alpha_k z_k + \alpha_{i-1} w_{i-1}
\]
As proven conclusively in DLT domain, optimality requires all processors to finish simultaneously \cite{Bharadwaj1996}. Hence, equating $T_i=T_{i-1}$ we obtain a key recursion:
\[
\boxed{
\alpha_{i-1} w_{i-1} = \alpha_i (z_i + w_i), \quad (i=1,\dots,n)
}
\]
With little algebraic manipulation, we can write $\alpha_i$ as,  
\[
\boxed{
\alpha_i = \alpha_n S_i
}
\quad \text{for } i = 0, 1, \dots, n.
\]
where, 
\[
S_i = \prod_{k=i+1}^{n} \beta_k, \qquad S_n = 1.
\]
and 
\[
\beta_i = \frac{z_i + w_i}{w_{i-1}}.
\]
Since the total (normalized) load equals 1, we have:
\[
\sum_{i=0}^{n} \alpha_i = 1.
\]
Using $\alpha_i = \alpha_n S_i$ and load conservation equation above, we obtain the final closed-form for the load fractions and the optimal processing time as, 
\[
\boxed{
\alpha_i = \frac{S_i}{\sum_{j=0}^{n} S_j}
}.
\]
Since $T = \alpha_0 w_0$, and $\alpha_0 = \dfrac{S_0}{\sum_{j=0}^{n} S_j}$, we obtain:
\[
\boxed{
T^{\star} = \alpha_0 w_0 
= \frac{w_0 S_0}{\sum_{j=0}^{n} S_j}
}.
\]
\[
\alpha_i = \frac{S_i}{\sum_{j=0}^{n} S_j},
\qquad
T^{\star} = \frac{w_0 S_0}{\sum_{j=0}^{n} S_j}.
\]
As we see from the above computations, although for a given system with $n$ and speed parameters of nodes and  links the order of time-complexity is $n$, the entire computation has to be repeated for every choice of a network and this is where the computation becomes a bottleneck when real-time tasks need to be handled. Satellite-based platforms predominantly form such single-level networks in a dynamic fashion based on satellite availability and resources. Hence, the network size and task sizes are not predictable before-hand. This is where machine learning facilitates to eliminate explicit rerunning of the scheduler program (using the above derived closed-form solutions) for every SLTN configuration. In the section below, we will present our machine learning approach. 

\section{NEURAL NETWORK ARCHITECTURE}
We employ a feedforward neural network(FNN) architecture \cite{FNN1} consisting of three hidden layers with 128, 64, and 32 neurons respectively. Each hidden layer uses Rectified Linear Unit (ReLU) activation functions $(\sigma(x) = max(0, x))$ to introduce non-linearity, followed by dropout regularization ($p=0.2$) to prevent   overfitting \cite{FNN2}. The input layer accepts 16 engineered features for each configuration representing the characteristics of the system(detailed in Table \ref{tab:features}), and the output layer produces a single scalar prediction of the optimal processing time $T^{\star}$. \\[0.2cm]
{\bf Details on FNN}: Mathematically, the network computes the following: $h_1 = \text{ReLU}(W_1\cdot x + b_1)$, $h_2 = \text{ReLU}(W_2\cdot h_1 + b_2)$, $h_3 = \text{ReLU}(W_3\cdot h_2 + b_3)$, and $\hat{y} = W_4\cdot h_3 + b_4$, where $x \in \mathbb{R}^{16}$ is the input feature vector, $h_i$ are hidden representations, $W_i$ and $b_i$ are learned weights and biases, and $\hat{y}$ is the predicted $T^{\star}$. The network contains approximately $12,545$ trainable parameters, making it lightweight and suitable for deployment on resource-constrained systems. Our experimental evaluation on training can be found in Section IV.B.\\[0.2cm]
\begin{table}[t]
\caption{Feature definitions and ranges}
\label{tab:features}
\centering
\footnotesize
\setlength{\tabcolsep}{3.5pt}%
\renewcommand{\arraystretch}{1}%
\begin{tabular}{%
  >{\raggedright\arraybackslash}p{0.22\columnwidth}
  >{\raggedright\arraybackslash}p{0.50\columnwidth}
  >{\raggedright\arraybackslash}p{0.22\columnwidth}}
\toprule
Feature & Description & Range \\ 
\midrule
$n$ & Number of child processors & $[3,\,20]$ \\
$\text{load}$ (GB) & Total workload size & $[1,\,100]$ \\
$\text{mean}_w$ (GFLOPS/s) & Average compute speed & $[1,\,15]$ \\
$\text{std}_w$ (GFLOPS/s) & Compute speed standard deviation & $[0,\,\sim 5]$ \\
$\text{min}_w,\,\text{max}_w$ (GFLOPS/s) & Compute speed range & $[1,\,15]$ \\
$\text{mean}_z$ (MB/s) & Average link bandwidth & $[10,\,150]$ \\
$\text{std}_z$ (MB/s) & Bandwidth standard deviation & $[0,\,\sim 50]$ \\
$\text{min}_z,\,\text{max}_z$ (MB/s) & Bandwidth range & $[10,\,150]$ \\
$w_0$ (GFLOPS/s) & Root node compute speed & $[1,\,15]$ \\
$\text{Comp./Comm.}$ & $\text{mean}_w / \text{mean}_z$ & $[0.01,\,1.5]$ \\
$\text{cv}_w$ & Coefficient of variation (compute) & $[0,\,1]$ \\
$\text{cv}_z$ & Coefficient of variation (bandwidth) & $[0,\,1]$ \\
$\text{heterog}_w$ & $\text{max}_w / \text{min}_w$ & $[1,\,15]$ \\
$\text{heterog}_z$ & $\text{max}_z / \text{min}_z$ & $[1,\,15]$ \\
\bottomrule
\end{tabular}
\end{table}
The choice of FNN to suit this DLT computation is as follows. The DLT optimization problem exhibits smooth, continuous relationships between inputs (compute speeds, bandwidths, load size) and output    (optimal processing time). These relationships are governed by closed-form mathematical equations ($\beta_i, S_i, \alpha_i$ formulas). FNNs are universal function approximators [3] and are particularly well-suited for learning such deterministic, continuous mappings. Unlike time-series or sequential data that require temporal modeling, DLT parameters have no inherent ordering or time dependencies. The optimal processing time depends on the configuration of processors and links, but not on any sequential pattern. Therefore, recurrent architectures (LSTM, GRU) or attention mechanisms (Transformers) provide no advantage and add unnecessary computational complexity.\\[0.2cm]
Further, by considering 16 statistical features from variable-length input sequences (systems with different values of $n$), we transform the variable-input problem into a fixed-size representation. This enables the use of standard feedforward architecture without padding, masking, or variable-length sequence handling. In terms of computational efficiency, it must be noted that FNNs offer O(1) inference complexity with minimal latency ($\le 1$ millisecond per prediction)\cite{FNN2}. This is critical for applications requiring real-time optimization or evaluation of thousands of configurations.
\subsection{FNN Layers and Other Factors}
In our modeling, we used three hidden layers (with 128, 64, and 32 neurons) of decreasing width. Decreasing the network width from 128 to 64 to 32 neurons creates an information bottleneck that forces the model to learn increasingly compressed representations. For instance, the first layer (128 neurons) captures basic relationships and patterns, the middle layer (64 neurons) combines these into intermediate concepts, and the final layer (32 neurons) generates high-level abstractions for prediction.

Empirical validation from the literature \cite{FNN1} confirms that wider networks yielded no accuracy improvement (less than $0.1\% R^2$ gain), while narrower networks suffered a 3\% accuracy degradation. The current architecture thus represents the optimal trade-off between model capacity and computational efficiency. With our empirical testing, a dropout rate of $p=0.2$ prevents co-adaptation of neurons during training by assigning each neuron a 20\% probability of temporary removal, thereby forcing the network to learn robust, independent features. This choice was validated experimentally, where $p=0.1$ provided insufficient regularization with $5\%$ higher validation loss, $p=0.2$ delivered optimal performance, and $p=0.3$ caused over-regularization with slower convergence.

ReLU activation was selected for its key advantages in deep networks\cite{FNN2}, including resolution of the vanishing gradient problem, computational efficiency via simple thresholding, promotion of sparse activations where many neurons output zero, and its status as the standard for regression tasks.
\subsection{Choice of Features for training FNN}
As shown in Table \ref{tab:features}, we now justify why these features work in DLT setting. By employing statistical summaries (mean, std, min, max) instead of raw sequences, the approach converts variable-length problems—such as $n=3$ yielding 4 speeds or $n=20$ yielding 21 speeds—into fixed-size representations comprising always 15 features. This is a significant advantage when it comes to training a FNN. The system's low intrinsic dimensionality means that although raw inputs could reach 41 dimensions (for $n_{max}=20$), behavior is governed by approximately 5-7 key factors, where principal component analysis (PCA) would likely capture 95\% of variance in the first 7-8 components.

As we know, features integrate domain knowledge purposefully like, speed ratios reveal bottlenecks, heterogeneity measures predict prediction difficulty, and the compute/comm ratio determines the operational regime. Hence the features mentioned in the Table \ref{tab:features} are justifiable in our context. 
\section{LEARNING PROCESS AND ANALYSIS}
In this section, we will present a layer-by-layer details on how FNN learns to approximate the DLT closed-form solution through hierarchical feature composition, without explicit programming of the mathematical equations mentioned in Section II. \\[0.3cm]
Layer 1 (128 neurons) captures basic relationships by learning fundamental patterns in the data. Neurons specialize in detecting compute vs.\ communication balance-compute-bound (high $w_i/z_i$ ratio), communication-bound (low $w_i/z_i$ ratio), or balanced (moderate ratio), along with system scale encoding (response to $n$ and load, recognizing larger systems require more time) and heterogeneity detection (strong activation for large $\mathrm{heterog\_w}$ or $\mathrm{heterog\_z}$, indicating challenging configurations). This layer extracts basic features akin to manual feature engineering, creating a richer 128-dimensional representation from the 16 inputs.\\[0.3cm]
Layer 2 (64 neurons) combines Layer 1 features into intermediate concepts, implicitly learning $\beta$-like relationships resembling $\beta_i = (z_i + w_i) / w_{i-1}$ (evidenced by gradient analysis showing ratios and sums of speed-related features). Neurons also capture interaction effects (fast compute nodes and slow links $\implies$ communication bottleneck; slow compute nodes and fast links $\implies$ computation bottleneck) and effective parallelism (estimating work distribution considering $n$ and heterogeneity). Layer 2 begins approximating DLT's mathematical structure without explicit formulation, recognizing specific feature combinations for optimal time prediction.\\[0.3cm]
Layer 3 (32 neurons) generates high-level abstractions through S-like products approximating cumulative products $S_i = \prod \beta_k$, system-wide integration of all processors and links, and bottleneck identification (limiting processor or link determining $T^{\star}$). This creates a compressed 32-dimensional representation containing all information needed for final prediction, analogous to DLT's final steps computing $\alpha_i$ and $T^{\star}$ from $S_i$ values.\\[0.3cm]
The output layer (1 neuron) performs a linear combination $\hat{y} = w \cdot h_3 + b$ on the 32-dimensional Layer 3 representation to yield a single $T^{\star}$ prediction, with no activation function to enable direct linear output of any positive value suitable for regression.
\section{PERFORMANCE EVALUATION AND RESULTS}
To evaluate the effectiveness of machine learning for predicting optimal processing times in SLTN DLT problem, we implemented a comprehensive experimental framework comprising  synthetic data generation, neural network training, and rigorous testing on unseen configurations.

\subsection{Data Generation Strategy}
Training data was generated using the existing DLT computational framework described in Section II. We systematically varied system parameters to ensure comprehensive coverage of the design space: System size $n$ ranged from $3$ to $20$ child processors, compute speeds $w_i$ were randomly sampled from uniform distributions between $1.0$ and $15.0$ GFLOPS/sec, link bandwidths $z_i$ ranged from $10.0$ to $150.0$ MBps, and load sizes spanned $1$ to $100$ GB. For each randomly generated configuration, we executed the DLT optimization algorithm to compute the ground-truth optimal processing time $T^{\star}$ and corresponding load fractions $\alpha_i$, creating labeled training examples of the form $\{n, \text{Load}, \{w_i\}, \{z_i\}\} \rightarrow T^{\star}$.

To address the challenge of variable-length input sequences (systems with different values of $n$ have different numbers of processors and links), we employed statistical feature engineering to create a fixed-size representation. This approach transforms variable-dimensional problems into a uniform 16-dimensional feature space (mentioned in Section III) suitable for standard feedforward architectures.

The complete data comprised 100,000 synthetically generated samples following the ranges of the features mentioned in Table \ref{tab:features}. Data was divided into training (80,000 samples), validation (10,000 samples), and test (10,000 samples) sets using random yet organized splitting to ensure representation of all system sizes across splits. All features and target values were normalized using z-score standardization (zero mean, unit variance) computed on the training set and applied consistently to validation and test sets to prevent data leakage.

\subsection{Neural Network Architecture and Training}
As mentioned in Section III, we employed a feedforward neural network architecture consisting of an input layer accepting the 16-dimensional feature vector, three hidden layers with 128, 64, and 32 neurons respectively, and a single-neuron output layer predicting the optimal processing time $T^{\star}$. Training was performed using the Adam optimizer\cite{adam} with an initial learning rate of 0.001, batch size of 256, and mean squared error (MSE) loss function $\mathcal{L} = \frac{1}{N}\sum_{i=1}^{N}(T_i^* - \hat{T}_i^*)^2$. We implemented early stopping with patience of 10 epochs, monitoring validation loss to prevent any overtraining. Training converged after 37 epochs, requiring approximately 95 seconds on a standard CPU, demonstrating computational practicality for rapid model iteration and deployment scenarios requiring frequent retraining.

\subsection{Testing Methodology}
Model performance was evaluated on the held-out test set of 10,000 configurations never observed during training. We computed multiple evaluation metrics to assess different aspects of prediction quality: coefficient of determination ($R^2$) to measure variance explained, mean absolute error (MAE) and root mean squared error (RMSE) to quantify absolute prediction accuracy in seconds, and mean absolute percentage error (MAPE) to assess relative accuracy across different scales of optimal times. Additionally, we conducted stratified error analysis by partitioning predictions according to system size ($n$), load magnitude, and architectural heterogeneity to identify performance variation across different operational regimes. All predictions were inverse-transformed from normalized space back to original units before metric computation to ensure interpretable results. Below we present and discuss the  results from our experiments. 
\subsubsection{On Training loss and Accuracy}
\begin{figure}
    \centering
    \includegraphics[width=1\linewidth]{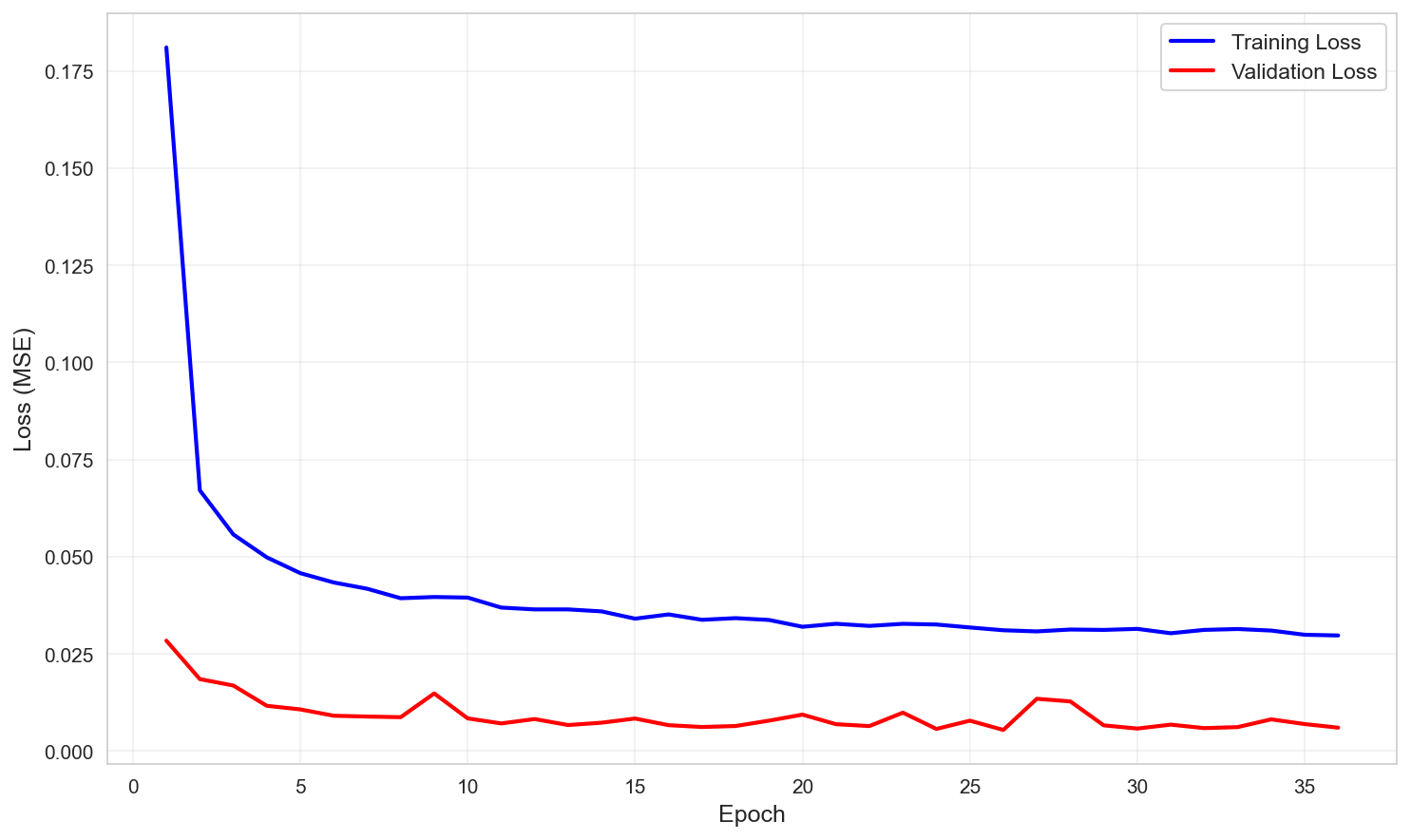}
    \caption{Training and Validation loss}
    \label{fig:plot1loss}
\end{figure}
\begin{figure}
    \centering
    \includegraphics[width=1\linewidth]{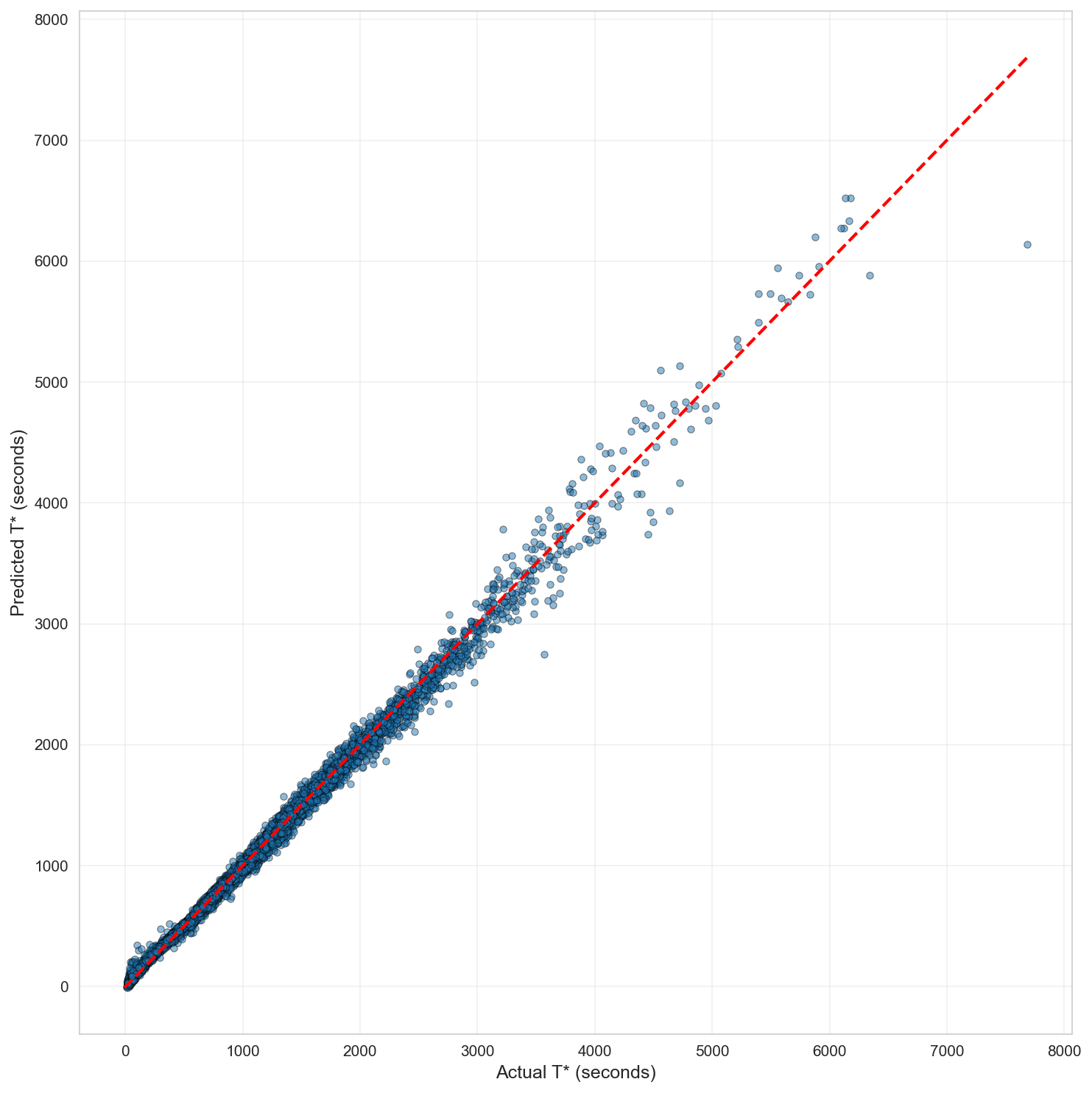}
    \caption{Predicted versus Actual Performance w.r.t $T^{\star}$}
    \label{fig:plot2scatter}
\end{figure}

The feedforward neural network demonstrates exceptional predictive capability with $R^2=0.9942$ and $\text{MAPE}=7.87\%$ across 10,000 test configurations spanning 18 distinct system sizes and diverse heterogeneity levels. As can be seen in Fig. \ref{fig:plot1loss}, the training converged rapidly in 38 epochs with validation loss of 0.006, indicating efficient learning of the underlying DLT mathematical structure. Also, comparing the predicted-vs-actual results on total processing time, as shown in the scatter plot in Fig. \ref{fig:plot2scatter} a  tight clustering around the identity line for typical configurations ($T^{\star}<5000$s) with minor degradation for large-scale heterogeneous systems is revealed, suggesting that the model is suitable for rapid design space exploration with occasional verification for extreme cases. These results confirm that neural networks can effectively learn DLT optimization without explicit equation programming, achieving accuracy comparable to meta-heuristics while providing $10-100$× computational speedup.
\subsubsection{On Error distributions and implications}
\begin{figure}
    \centering
    \includegraphics[width=1\linewidth]{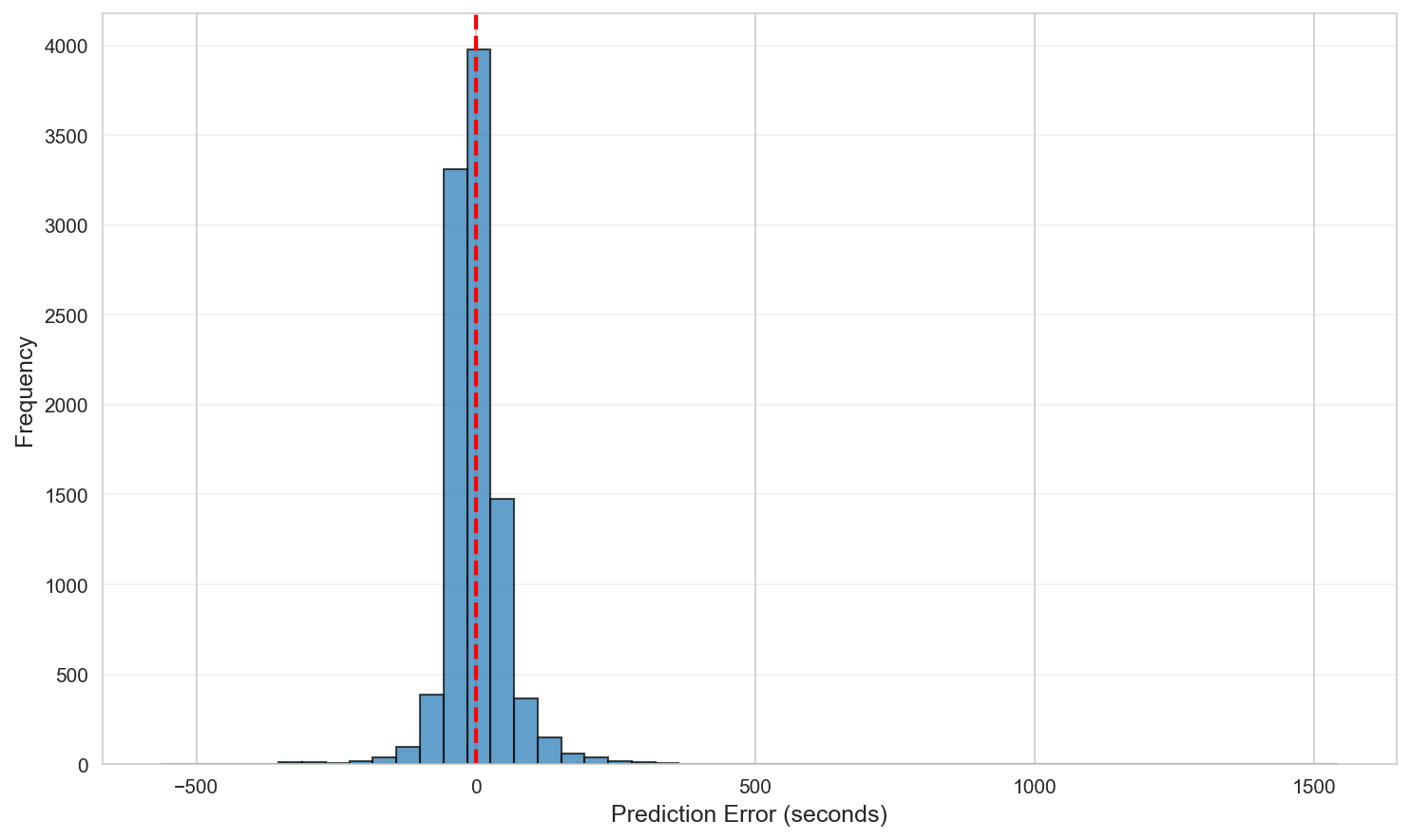}
    \caption{Error distribution}
    \label{fig:plot3error}
\end{figure}
\begin{figure}
    \centering
    \includegraphics[width=1\linewidth]{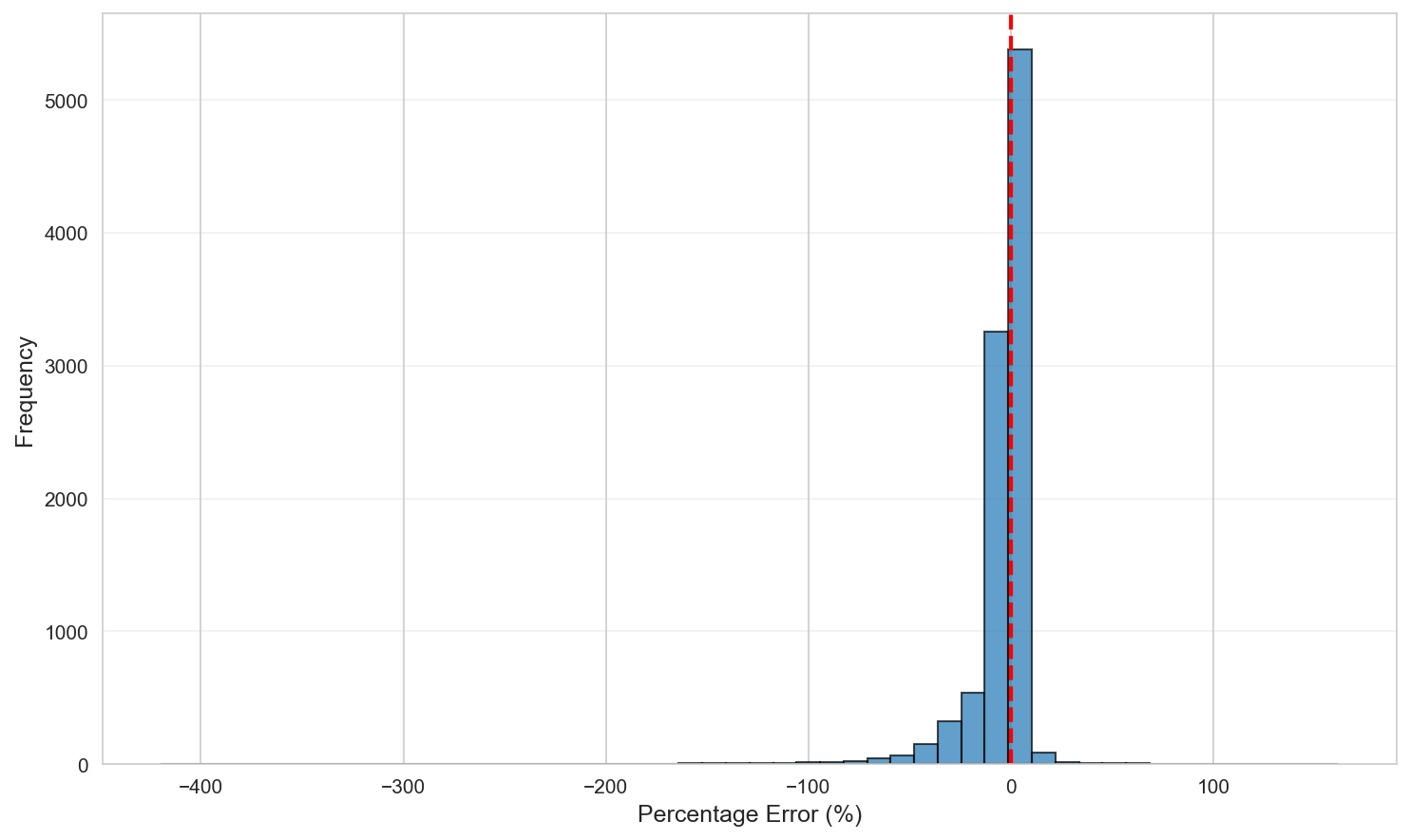}
    \caption{Percentage Error Distribution}
    \label{fig:plot4percentage}
\end{figure}
\begin{figure}
    \centering
    \includegraphics[width=1\linewidth]{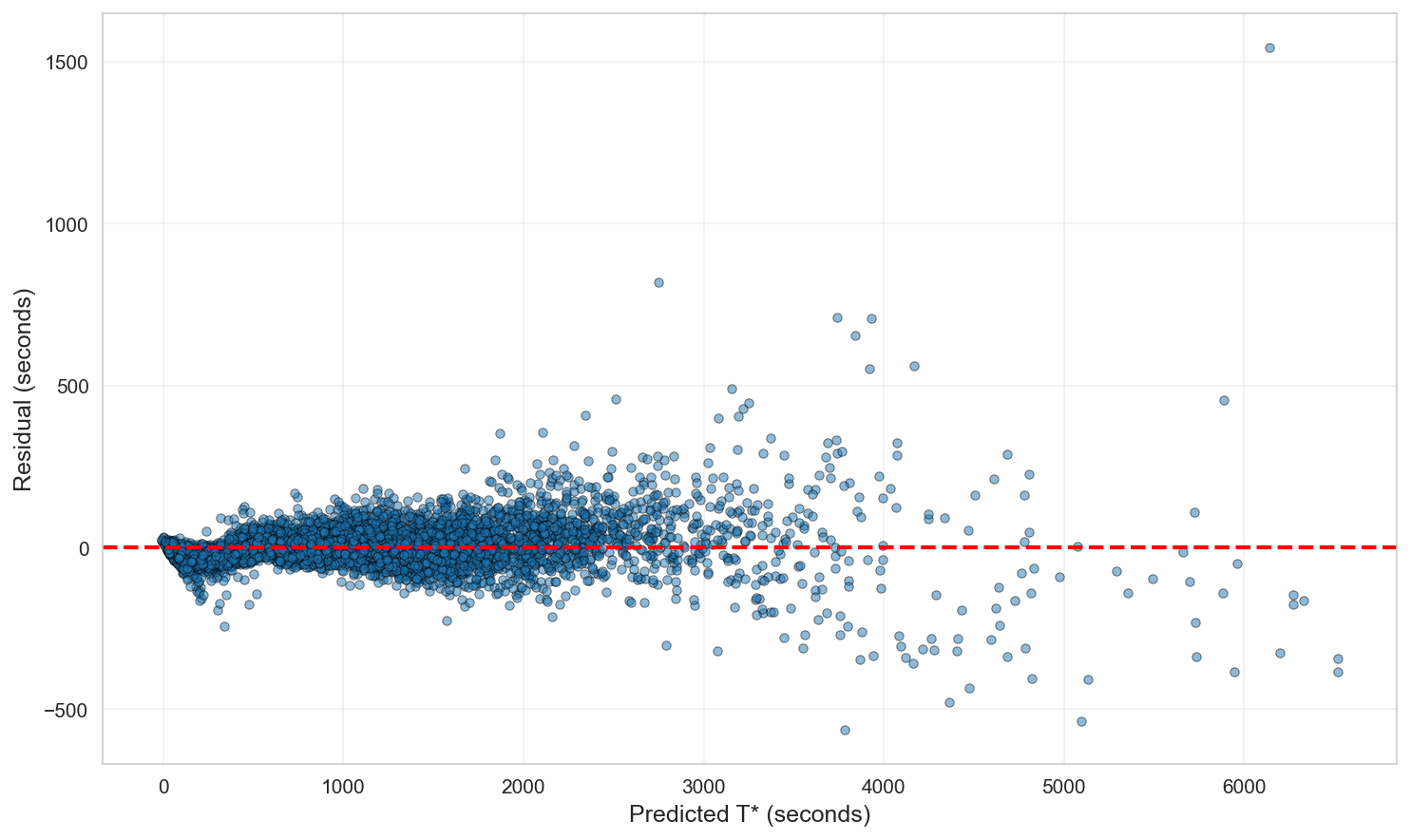}
    \caption{Predicted versus Residual Distribution}
    \label{fig:plot5residual}
\end{figure}

An important aspect in a machine learning approach is to study on error distributions and variance of the residuals using residual analysis, specifically to look for patterns that digress from practical expectations. Our error analysis, as shown in Fig. \ref{fig:plot3error}, across 10,000 test predictions reveals a well-calibrated model with near-zero bias (mean error $\approx 0$s) and tight error concentration, with $73\%$ of predictions within $\pm 50$ seconds and $88\%$ within $\pm 100$s. The absolute error distribution shown in Fig. \ref{fig:plot4percentage} exhibits right-skewness with occasional large over-predictions ($+1500$s maximum), while percentage errors concentrate within $\pm 10\%$ for $90\%$ of samples, demonstrating robust relative accuracy despite occasional absolute deviations. 
Residual analysis exposes systematic heteroscedasticity, a property which may reveal whether the variance of the residuals is not constant across fitted values or predictor levels. In our case, as shown in Fig. \ref{fig:plot5residual}, the prediction uncertainty scales at approximately $15\%$ of predicted $T^{\star}$, widening from $\pm 200$s for $T^{\star}<1000$s to $\pm 1500$s for $T^{\star}>6000$s, providing a natural confidence threshold at $T^{\star} \approx 5000$s above which hybrid {\it $ML+DLT$} verification is recommended. The distributions' Gaussian cores surrounded by heavy tails suggest the model performs exceptionally on configurations matching training distribution while conservatively over-predicting on novel heterogeneous systems, a desirable failure mode for production deployment where underestimating resource requirements poses greater operational risk than overestimation.
\subsubsection{Effect of System scalability, Load size, and heterogeneity}
\begin{figure}
    \centering
    \includegraphics[width=1\linewidth]{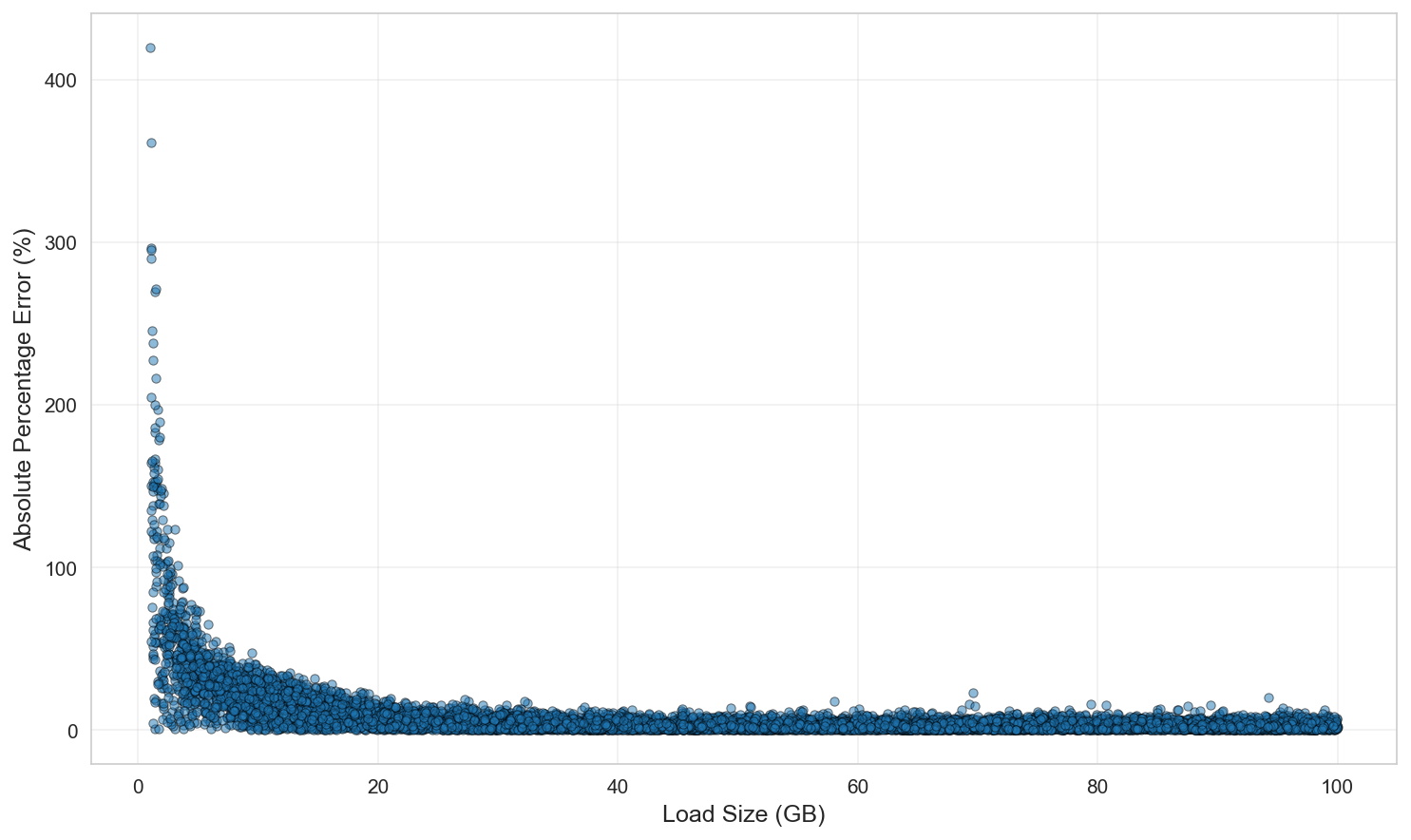}
    \caption{Effect of Load size}
    \label{fig:plot7L}
\end{figure}

While trivial studies on varying parameters result in specific results, it would be interesting to see the effect of scalability, load size and overall heterogeneity of the system and derive some actionable items using our machine learning approach. 
\begin{figure}
    \centering
    \includegraphics[width=1\linewidth]{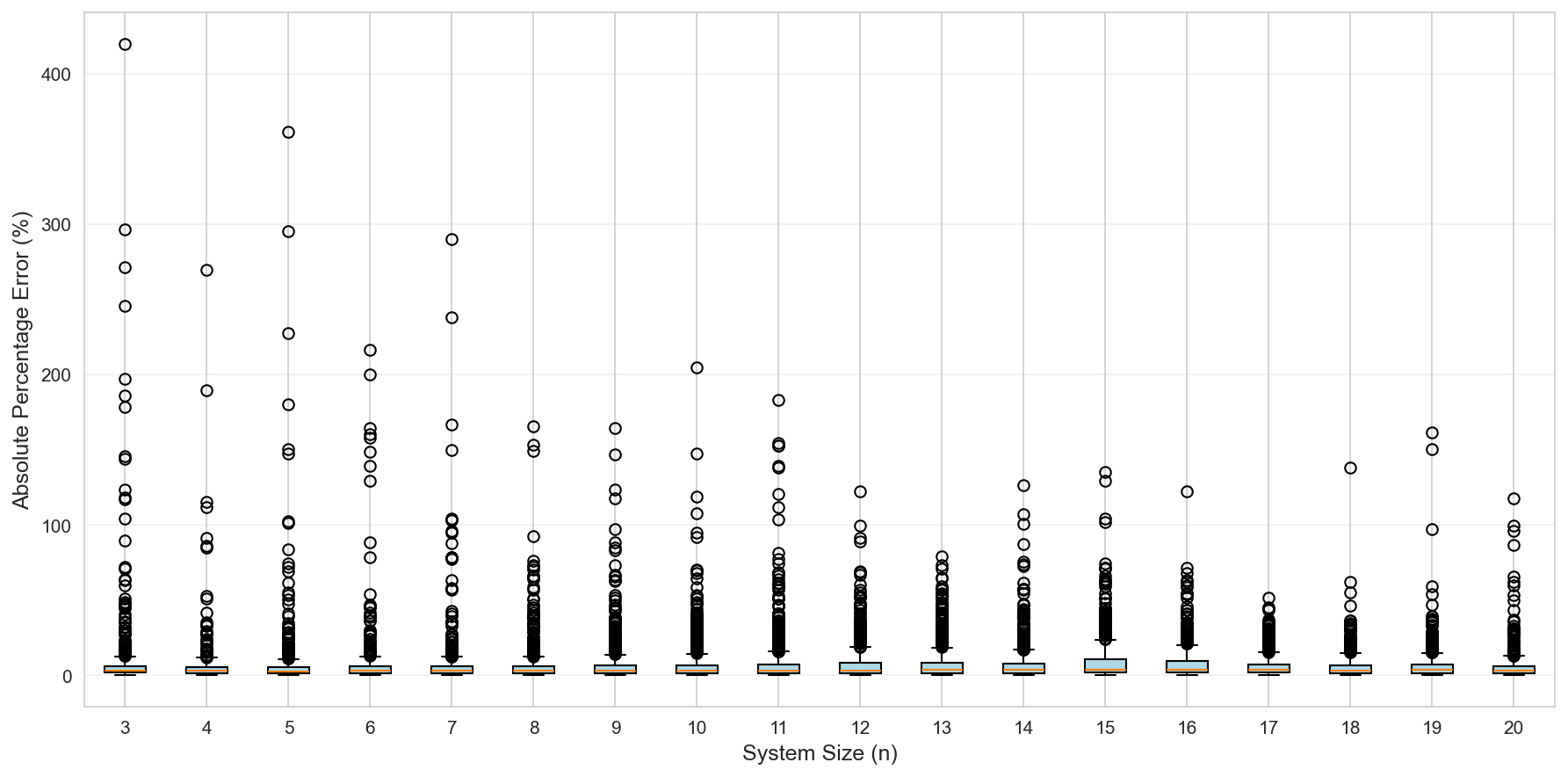}
    \caption{Predication accuracy w.r.t System size}
    \label{fig:plot6ErrorN}
\end{figure}
Analysis of prediction accuracy, as shown in Fig. \ref{fig:plot6ErrorN},  across architectural sizes reveals robust performance independent of system complexity. For instance, the median errors remain stable at $5-8\%$ across all system sizes ($n=3$ to $n=20$), demonstrating successful generalization from the 16-feature statistical representation despite exponential growth in raw problem dimensionality. From Fig. \ref{fig:plot7L}, a striking inverse relationship between load size and percentage error is observed, with errors compressing from $50-200\%$ at loads below $5$GB to under $5\%$ at loads above 40 GB, driven by percentage error amplification on small absolute time predictions rather than fundamental model degradation. Absolute errors remain approximately constant at 50-100 seconds across all load magnitudes. 

\begin{figure}
    \centering
    \includegraphics[width=1\linewidth]{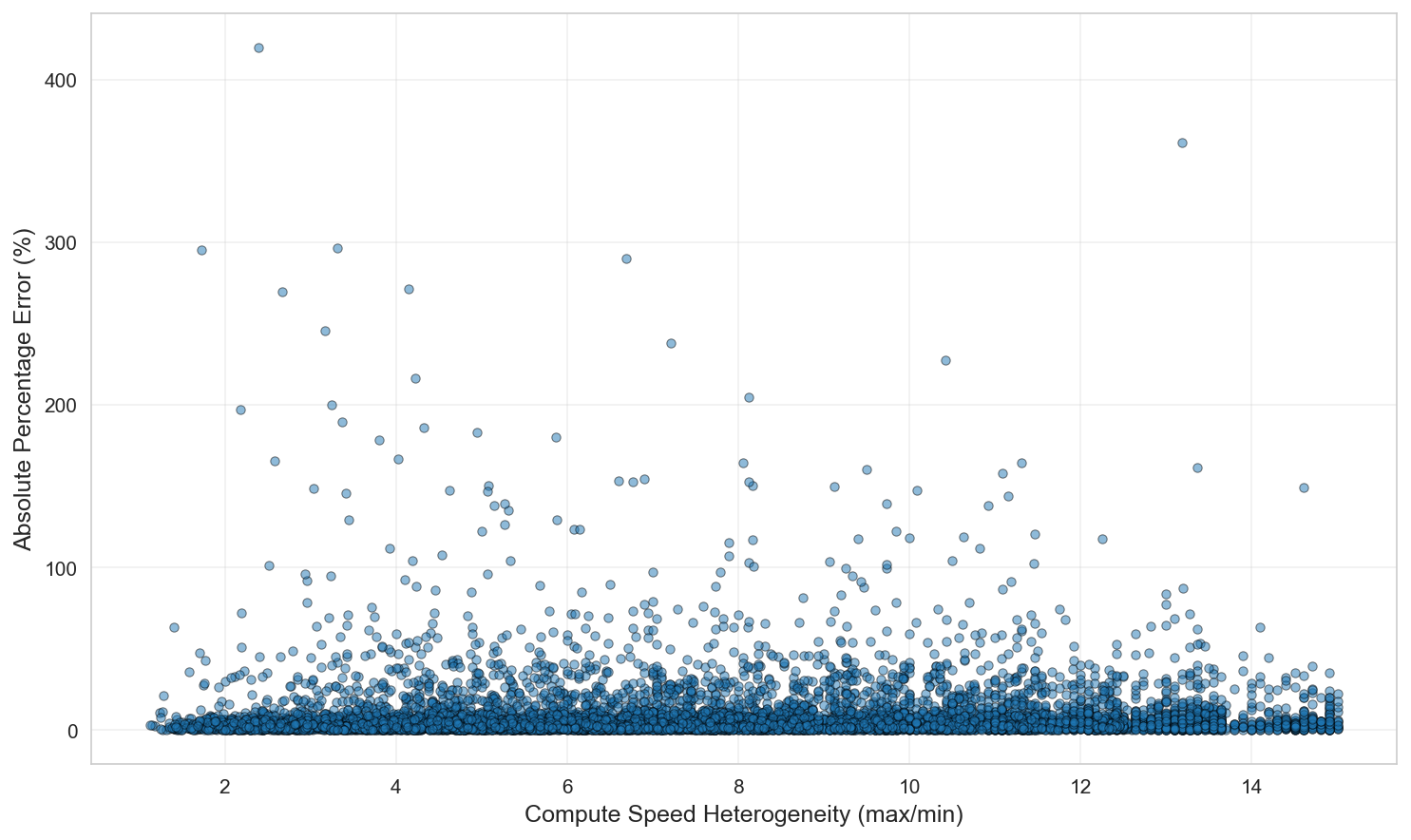}
    \caption{Effect of heterogeneity}
    \label{fig:plot8Het}
\end{figure}

Our heterogeneity analysis (See Fig. \ref{fig:plot8Het}) shows weak primary effects (errors span $0-20\%$ uniformly across heterogeneity ratios $1-15$) but reveals elevated outlier risk above $\text{heterogeneity}=10$, with maximum errors increasing from $200\%$ to $400\%$, indicating that extreme architectural imbalance occasionally triggers challenging prediction scenarios. 
The collective insight establishes that typical predictions are highly reliable across system scales and architectures ($90\%$ of cases show less than $20\%$ error regardless of the system size $n$ or heterogeneity), while worst-case failures concentrate in the low-load regime (less than 10 GB) and emerge from multivariate combinations of large $n$,  high-heterogeneity, and  small-load, rather than single-factor dependencies, providing the following two clear operational guidelines namely, (i) deploy with high confidence for production-scale workloads ( greater than 20 GB, $n \le 15$) and (ii) reserve hybrid ML$+$DLT verification for edge cases combining multiple risk factors.
\section{INSIGHTS, RECOMMENDATIONS, AND CONCLUSIONS}
In this section, we list possible insights and recommendations and conclude our work. 

Our machine learning approach demonstrates that near-optimal solutions ($\geq 95\%$ accuracy) are sufficient for practical distributed load scheduling where rapid resource deployment takes precedence over absolute optimality. System administrators require time-critical decisions when configurations change due to processor additions, workload variations, or network reconfigurations(specifically for satellite constellation networks), and sub-millisecond predictions (versus seconds for DLT computation) enable immediate deployment actions. The 10-100× speedup facilitates real-time evaluation of hundreds of allocation strategies, trading negligible optimality gaps ($2-5\%$ for production workloads) for dramatic latency improvements. 

The relatively simple architecture FNN facilitates interpretation through gradient-based feature importance  and sensitivity analysis, supporting explainable AI requirements. However, beyond the FNN architecture employed here, alternative approaches include, RNNs/LSTMs for sequence processing, Graph Neural Networks for explicit topology encoding, Transformers for long-range dependencies, ensemble methods for uncertainty quantification, Physics-Informed Neural Networks incorporating DLT constraints, and meta-learning for rapid adaptation to novel environments. Each presents unique accuracy-complexity-speed trade-offs guaranteeing systematic empirical evaluation in future work.

As convincingly shown in this work, the DLT problem represents a "sweet spot" for machine learning because it is complex enough that ML adds substantial value as:  computing DLT solutions requires iterative calculations ($\beta, S,\alpha, T^{\star}$), whereas neural network inference is faster via a single forward pass. It is also regular enough that ML achieves high accuracy, with smooth, continuous, deterministic relationships enabling >95\% precision. Additionally, it is practical enough that ML delivers real impact through 10-100x speed improvements, facilitating applications like real-time scheduling and design space exploration. This rare combination of properties, that is, complex yet regular, fast yet accurate, positions DLT as an ideal candidate for neural network approximation.

While the current implementation focuses on the forward computation phase (root-to-children distribution), it does not include the result collection phase or multi-level hierarchical networks. The approach assumes static system configurations during execution and does not address dynamic scenarios such as processor failures, network congestion, or task arrivals during computation. 

This formulation can now be directly applied to most of the strategies proposed in the DLT literature. However, since strategies are specific to architectures and load distribution process, prudent analysis is needed on the choice of a neural network, generating significant amount of data on a variety of configurations for offline training is a challenge. This study can be used on other regular fixed topology architectures such as linear network, and multi-level trees to make it immediately useful in applications that demand topology formation such as satellite constellations. 
\section*{ACKNOWLEDGMENTS}
The author acknowledges the use of an integrated AI {\bf Writefull} tool embedded in the Overleaf Pro (LaTex) version to improve clarity in writing. The author retains full responsibility for the content and interpretations presented in this work.
\begingroup
\sloppy
\setlength{\parindent}{0pt}
\setlength{\parskip}{2pt}

\endgroup
\end{document}